\documentclass[conference]{IEEEtran}
\IEEEoverridecommandlockouts

\usepackage{cite}
\usepackage{amsmath,amssymb,amsfonts}
\usepackage{algorithmic}
\usepackage{graphicx}
\usepackage{textcomp}
\usepackage{xcolor}
\usepackage{makecell}
\def\BibTeX{{\rm B\kern-.05em{\sc i\kern-.025em b}\kern-.08em
    T\kern-.1667em\lower.7ex\hbox{E}\kern-.125emX}}
\begin{document}

\title{Knowledge-Guided Prompt Learning for Deepfake Facial Image Detection}

\author{
\IEEEauthorblockN{Hao Wang$^{\dagger}$, Cheng Deng$^{\ddagger}$, Zhidong Zhao$^{\dagger\star}$}
\IEEEauthorblockA{$^{\dagger}$ School of Cyberspace Security, Hangzhou Dianzi University, Hangzhou, China\\
$^{\ddagger}$ School of Electronic Engineering, Xidian University, Xi'an, China
}
\thanks{$\star$This work was supported in part by the Zhejiang Provincial Natural Science Foundation of China under Grant No.LQN25F020014, in part by the Fundamental Research Funds for the Provincial Universities of Zhejiang under Grant No.GK249909299001-030, in part by the Zhejiang Provincial Natural Science Foundation of China under Grant No.LDT23F01012F01, and in part by the Zhejiang Province Key R\&D Program Project under Grant No.2024C01102. Zhidong Zhao is the corresponding author.}
}

\maketitle

\begin{abstract}
Recent generative models demonstrate impressive performance on synthesizing photographic images, which makes humans hardly to distinguish them from pristine ones, especially on realistic-looking synthetic facial images. Previous works mostly focus on mining discriminative artifacts from vast amount of visual data. However, they usually lack the exploration of prior knowledge and rarely pay attention to the domain shift between training categories (e.g., natural and indoor objects) and testing ones (e.g., fine-grained human facial images), resulting in unsatisfactory detection performance. To address these issues, we propose a novel knowledge-guided prompt learning method for deepfake facial image detection. Specifically, we retrieve forgery-related prompts from large language models as expert knowledge to guide the optimization of learnable prompts. Besides, we elaborate test-time prompt tuning to alleviate the domain shift, achieving significant performance improvement and boosting the application in real-world scenarios. Extensive experiments on DeepFakeFaceForensics dataset show that our proposed approach notably outperforms state-of-the-art methods.
\end{abstract}

\begin{IEEEkeywords}
deepfake facial image detection, knowledge-guided prompt learning, domain shift, test-time prompt tuning.
\end{IEEEkeywords}

\section{Introduction}
With the rapid development of deep generative models, including Generative Adversarial Networks (GANs) \cite{goodfellow2020generative} and diffusion-based models \cite{ho2020denoising}, the creation of realistic synthetic content has achieved remarkable progress, especially on human facial images. However, leveraging these AI-generated facial images poses a significant risk for security in various fields such as social media, politics and economy. To fight against this risk, a variety of detection approaches have been proposed to differentiate whether an image is real or fake.

\begin{figure}[t]
	\centering
	\includegraphics[scale=0.4]{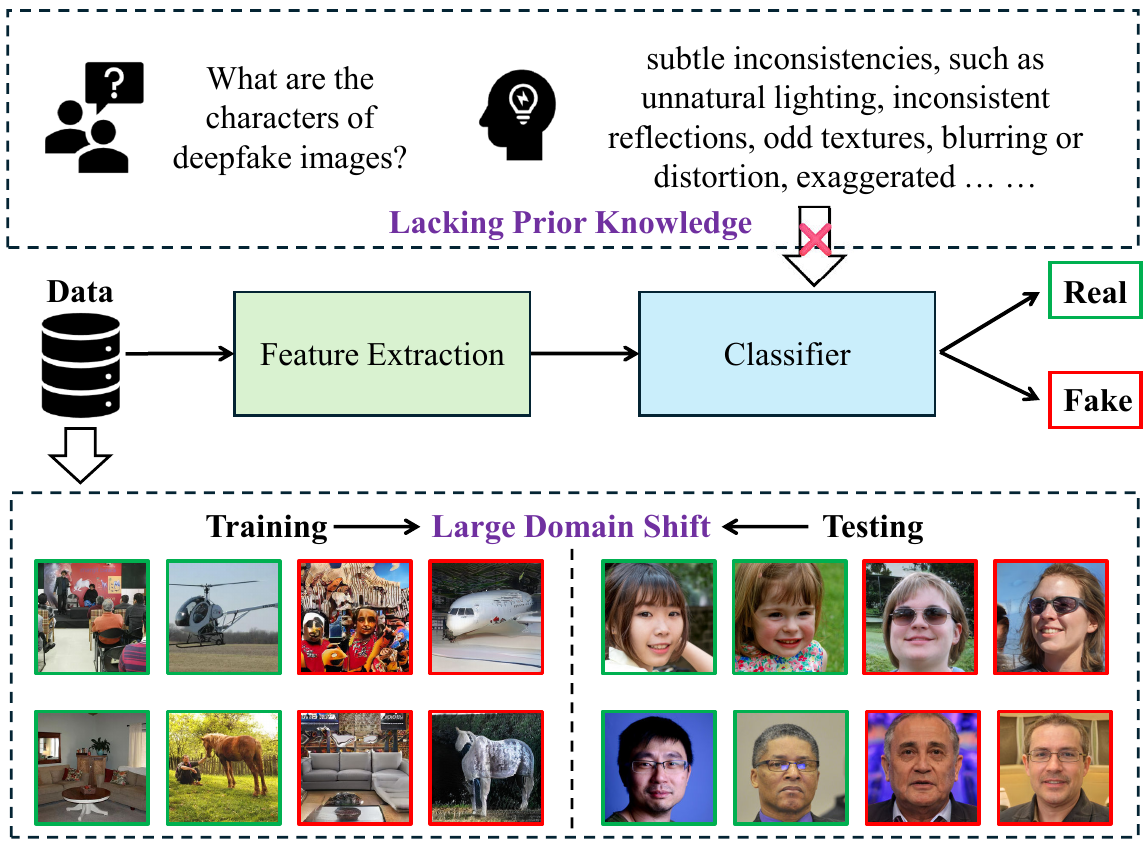}
	\caption{Existing approaches mostly focus on forgery artifacts extraction. However, they lack the exploration of expertise and pay little attention to the domain shift between training categories and testing ones.} \label{motivation}
\end{figure}

Early methods \cite{wang2020cnn,yu2019attributing,marra2018detection,chai2020makes,schwarcz2021finding,zhao2021multi,yu2022patch,mandelli2022detecting,frank2020leveraging,qian2020thinking,doloriel2024frequency} can be roughly divided into two categories, i.e., learning on spatial and frequency domain, respectively. The former methods extract global semantic features \cite{wang2020cnn} \cite{yu2019attributing} \cite{marra2018detection} or local artifacts \cite{chai2020makes} \cite{schwarcz2021finding} \cite{zhao2021multi} \cite{yu2022patch} \cite{mandelli2022detecting} for detection. The latter ones identify forgery clues via frequency-domain analysis, such as 2d-DCT \cite{frank2020leveraging} \cite{qian2020thinking} and 2d-FFT \cite{doloriel2024frequency}. To utilize various complementary features for better detection, several approaches attempt to integrate multiple kinds of features. For example, GLFF \cite{ju2023glff} fuses global and local semantic features to mine rich and discriminative representations in spatial domain. Tian \textit{et al.} \cite{tian2023frequency} aggregate diverse information from both spatial and frequency domain to alleviate overfitting issues. Recently, vision-language models have attracted ever-increasing attention in computer vision community. For deepfake image detection, UnivFD \cite{ojha2023towards} utilizes fixed visual feature space of large pre-trained model (i.e., CLIP \cite{radford2021learning}) to achieve universal fake image detection. FatFormer \cite{liu2024forgery} adapts the pre-trained vision-language space via frequency feature mining and textual prompts learning, leading to nontrivial generalization improvement. However, there are two significant drawbacks among these methods, as illustrated in Fig. \ref{motivation}.

\begin{figure*}[t]
	\centering
	\includegraphics[scale=0.63]{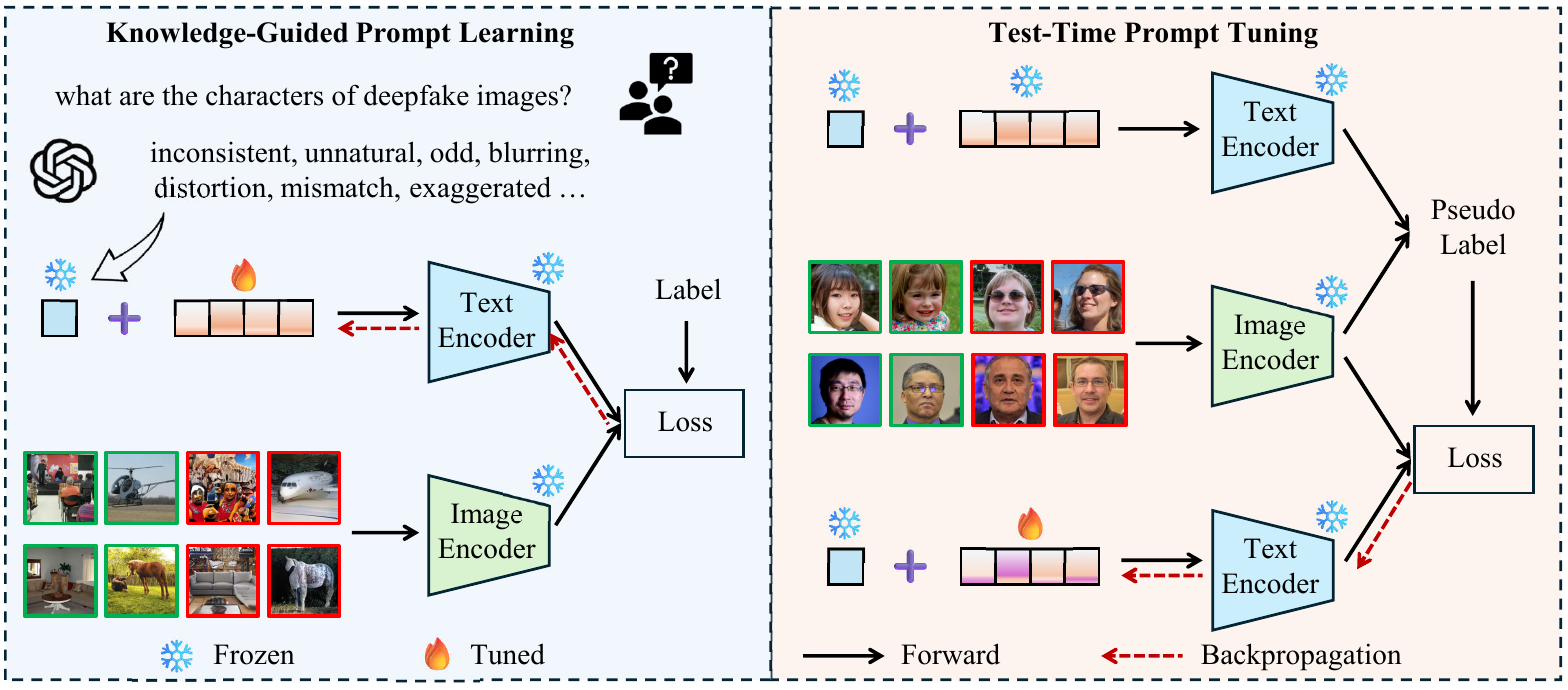}
	\caption{The overall framework, which consists of knowledge-guided prompt learning and test-time prompt tuning. The former elicit prior knowledge from a large language model to construct more meaningful prompts. The latter obtains pseudo label for testing data and then tunes prompt to alleviate domain shift.} \label{framework}
\end{figure*}

Firstly, they lack the exploration and incorporation of prior knowledge to differentiate whether an image is real or fake, resulting in unsatisfactory detection performance. Specifically, earlier works mostly learn the decision boundary among visual samples, without the participation of prior knowledge. FatFormer, the more recent vision-language model, merely utilizes learnable prompts along with simple category names, i.e., `real' and `synthetic'. Research \cite{zhou2023anomalyclip} \cite{li2024promptad} show that augmenting the prompts with relevant semantic concepts (i.e., prior knowledge) can remarkably boost the performance. Unfortunately, the augmentation can be hardly achieved without the help of expert-level annotation of dataset. To address this issue, we retrieve meaningful concepts from a large language model such as GPT4 \cite{achiam2023gpt} to construct prompts, improving both detection performance and the interpretability of prompts.

Secondly, most approaches ignore the large domain shift between training categories (e.g., natural and indoor objects) and testing ones (e.g., fine-grained human facial images). Concretely, the training process is conducted on easily distinguished synthetic images, such as the `person' and `horse' illustrated in Fig. \ref{motivation}, while the testing is evaluated on hardly differentiated forgery facial images. This shift will make the classifier to recognize some realistic-looking fake images as pristine ones, leading to unsatisfactory detection performance and hindering the application in real-word scenarios. To address this issue, we develop test-time prompt tuning to further improve detection performance, with the strict condition that no ground-truth label is provided.

The main contributions of this work are as follows:
\begin{itemize}
	\item We propose a novel knowledge-guided prompt learning method by incorporating prior knowledge from a large language model for deepfake facial image detection;
	\item We devise a simple yet effective test-time prompt tuning method to alleviate the large domain shift between training categories and testing ones;
	\item Extensive experiments on challenging DeepFakeFaceForensics dataset demonstrate that our proposed approach significantly outperforms state-of-the-art methods.
\end{itemize}

\section{Methodology}
The overall structure of our proposed method is illustrated in Fig. \ref{framework}, consisting of Knowledge-Guided Prompt learning (KGP) and Test-Time Prompt tuning (TTP). To incorporate expert-level knowledge for detection, we retrieve forgery-related concepts from an off-the-shelf large language model, avoiding professional and labor-intensive annotation of dataset. Moreover, to mitigate the large domain shift between training categories and testing ones, we take the model optimized on training samples as reliable annotator to obtain pseudo labels of partial testing data. Then the prompts are further tuned with these pseudo supervision signals.

\subsection{Knowledge-Guided Prompt Learning}
Traditional methods \cite{zhou2022conditional} \cite{khattak2023maple} optimize learnable prompts with the guidance of category names, e.g., airplane, person and so on. For deepfake image detection, they are commonly restricted as `real' and `fake' \cite{laiti2024conditioned} \cite{liu2024forgery} or it is just replaced as `is this photo real' \cite{chang2023antifakeprompt} for prompt learning. However, research \cite{li2024promptad} demonstrates that simple concepts can not adequately unleash the potential of vision-language models, highlighting the importance of incorporating expert-level knowledge.

\begin{table*}[t]
	\centering
	\caption{Evaluation results (AUC) on DeepFakeFaceForensics dataset. Following GLFF \cite{ju2023glff}, the AUC of each subset is the average AUC among 6 generative models.}
	\label{aucsota}
	\begin{tabular}{c|c|c|c|c|c|c|c}
		\hline
		\textbf{Testing Subset} & \textbf{CNN-aug \cite{wang2020cnn}} & \textbf{GAN-DCT \cite{frank2020leveraging}} & \textbf{Nodown \cite{gragnaniello2021gan}} & \textbf{BeyondtheSpectrum \cite{he2021beyond}} & \textbf{PSM \cite{ju2022fusing}} & \textbf{GLFF \cite{ju2023glff}} & \textbf{Ours} \\ \hline
		\textbf{Unprocessed} & 0.723 & 0.656 & \textbf{0.970} & 0.819 & 0.901 & 0.906 & 0.939 \\ \hline
		\textbf{Common Post-processing} & 0.710 & 0.443 & 0.823 & 0.624 & 0.878 & 0.887 & \textbf{0.938} \\ \hline
		\textbf{Face Blending} & 0.795 & 0.483 & 0.888 & 0.558 & 0.877 & \textbf{0.905} & 0.771 \\ \hline
		\textbf{Anti-forensics} & 0.605 & 0.504 & 0.894 & 0.644 & 0.863 & 0.834 & \textbf{0.963} \\ \hline
		\textbf{Multi-image Compression} & 0.217 & 0.646 & 0.105 & 0.577 & 0.411 & 0.547 & \textbf{0.889} \\ \hline
		\textbf{Mixed} & 0.528 & 0.497 & 0.468 & 0.470 & 0.724 & 0.801 & \textbf{0.970} \\ \hline
		\textbf{Average} & 0.596 & 0.538 & 0.691 & 0.616 & 0.775 & 0.813 & \textbf{0.911} \\ \hline
	\end{tabular}
\end{table*}

Hence, we retrieve forgery-related concepts from a large language model to construct more meaningful prompts, without the help of professional and labor-intensive annotation. Specifically, we provide a query such as \textit{`what are the characters of deepfake images?'} for large language model and collect the forgery-related concepts, e.g., `unnatural', 'inconsistent' and 'blurred'. Formally, the prompts can be defined as follows:
\begin{align}
	\centering
	P_{real} & = [P_1] [P_2] \cdots [P_N] [real], \\
	P_{fake}^i & = [P_1] [P_2] \cdots [P_N] [\phi^i],
\end{align}
where $P_{real}$ and $P_{fake}^i$ denote the prompts for pristine images and synthetic ones, respectively. $\phi^i$ indicates the $i$-th forgery-related concept of $\Phi = \{fake, blurred, \cdots, unrealistic\}$, which is retrieved from large language model. $N$ represents the number of learnable prompts.

Since there are multiple prompts for synthetic image, we calculate its prototype to avoid multiple similarity computation, which can be described as follows:
\begin{align}
	G_{real} & = G(P_{real}), \\
	G_{fake} & = \frac{1}{|\Phi|} \sum_{i=1}^{|\Phi|} G(P_{fake}^i),
\end{align}
where $G_{real}$ and $G_{fake}$ are textual features for pristine prompt and synthetic one, respectively. $G$ denotes the text encoder of pre-trained vision-language model. Mathematically, given input images $X$ and their ground-truth labels $Y$ (i.e., $0$ for real and $1$ for fake), the loss can be formulated as follows:
\begin{align}
	\mathcal{L}  = & -Y log \delta_{fake}  - (1 - Y) log \delta_{real}, \\
	\delta_{fake}  = & \frac{exp(S_{fake} / \tau)}{exp(S_{real} / \tau) + exp(S_{fake} / \tau)}, \\
	\delta_{real}  = & \frac{exp(S_{real} / \tau)}{exp(S_{real} / \tau) + exp(S_{fake} / \tau)}, \\
	S_{fake} = cos(F(&X), G_{fake}), S_{real} = cos(F(X), G_{real}),
\end{align}
where $\delta_{fake}$ and $\delta_{real}$ represent the predicted probabilities of images $X$ being synthetic or pristine, respectively. $S_{fake}$ and $S_{real}$ denote the similarity scores calculated with cosine function $cos$. Besides, $F$ and $\tau$ are image encoder and temperature coefficient of pre-trained vision-language model, e.g., CLIP.

\subsection{Test-Time Prompt Tuning}
After knowledge-guided prompt learning, the model has a certain capability to differentiate between pristine and synthetic images. However, the significant domain shift between the training and testing categories makes the model prone to misidentifying synthetic images as pristine, hindering the application in real-world scenarios.

To address the domain shift, we tune the prompts on testing data, leveraging pseudo labels generated from the model trained with knowledge-prompt learning. Specifically, we calculate the similarity scores and sort them in descending order. Given testing set $X = \{x_1, x_2, \cdots, x_M\}$ with total $M$ samples, the selection can be defined as follows:
\begin{align}
	X_{real} & = \{x_i\ |\ \delta_{real}^i > T_{real}\}, \\
	X_{fake} & = \{x_i\ |\ \delta_{fake}^i > T_{fake}\}, \\
	| X_{real} | &\le TopK, \ | X_{fake} | \le TopK, 
\end{align}
where $\delta_{fake}^i$ and $\delta_{real}^i$ represent the predicted probabilities of image $x_i$ being synthetic or pristine, respectively. $T_{real}$ and $T_{fake}$ are thresholds for selection of pristine and synthetic images. $TopK$ is the largest number of samples to be selected. Then the training set can be described as follows:
\begin{equation}
	\widehat{X} = \{X_{real}, \ X_{fake}\}, \ \widehat{Y} \in \{0, 1\},
\end{equation}
where $\widehat{y}_i$ is $0$ if $x_i \in X_{real}$ otherwise $1$. Given samples $\widehat{X}$ and its pseudo labels $\widehat{Y}$, the loss for test-time prompt tuning can be formulated as follows:
\begin{equation}
	\mathcal{L} = -\widehat{Y} log \delta_{fake}  - (1 - \widehat{Y}) log \delta_{real},
\end{equation}
where $\delta_{fake}$ and $\delta_{real}$ represent the predicted probabilities of images $\widehat{X}$ being synthetic or pristine, respectively.

\begin{table}[t]
	\centering
	\caption{Evaluation results (OA) on DeepFakeFaceForensics dataset. `-' denotes that result is not reported in original paper.}
	\label{oasota}
	\begin{tabular}{c|c|c}
		\hline
		\textbf{Test Data} & \textbf{GLFF \cite{ju2023glff}} & \textbf{Ours} \\ \hline
		\textbf{Unprocessed} & - & 0.826 \\ \hline
		\textbf{Common Post-processing} & - & 0.795 \\ \hline
		\textbf{Face Blending} & - & 0.548 \\ \hline
		\textbf{Anti-forensics} & - & 0.887 \\ \hline
		\textbf{Multi-image Compression} & - & 0.738 \\ \hline
		\textbf{Mixed} & 0.355 & 0.854 \\ \hline
		\textbf{Average} & - & 0.774 \\ \hline
	\end{tabular}
\end{table}

\section{Experiment}
\subsection{Dataset}
The training dataset we employed is the same as that in \cite{wang2020cnn} \cite{frank2020leveraging} \cite{gragnaniello2021gan} \cite{he2021beyond} \cite{ju2022fusing} \cite{ju2023glff}. It comprises around 360K pristine and 360K synthetic images, where the former are from LSUN dataset \cite{yu2015lsun} with 20 categories (e.g., airplane, cat, horse, soft, chair, person, etc.) and the latter are generated by ProGAN \cite{karras2017progressive} model with same categories.

We utilize DeepFakeFaceForensics \cite{ju2023glff} as testing dataset to evaluate the detection performance. It is a highly-diverse synthesized deepfake facial image dataset, which considers 6 generative models (e.g., GAN-based, transformer-based, diffusion-based, etc.) and 5 post-processing operations (e.g., compression, blurring, manipulation, anti-forensics, multi-image compression, etc.) to approach the real-world applications.

For evaluation, we adopt common metrics, including Overall Accuracy (OA) and Area Under ROC Curve (AUC), to demonstrate the detection performance. To be noted, accuracy is measured under the threshold $0.5$ across all experiments.

\subsection{Experiment Setup}
We use the text and image encoder from pre-trained CLIP model (i.e., ViT-L/14 variant \cite{dosovitskiy2020image}) to extract textual and visual features. Note that their parameters are frozen all the time. The number of learnable prompts is $1$. The batch size is 256 and 128 for training and testing respectively, as the same as in \cite{ojha2023towards}. The thresholds $T_{real}$ and $T_{fake}$ are set as $0.999$ and $0.5$ along with $TopK = 128$ for selection. We adopt Adam \cite{kingma2014adam} optimizer with learning rate as $1e-4$ during knowledge-guided prompt learning stage while $5e-5$ at test-time prompt tuning stage. The data augmentation strategies are the same as that in \cite{ju2023glff} during training.

\begin{table}[t]
	\centering
	\caption{Ablation studies on DeepFakeFaceForensics dataset.}
	\label{ablation}
	\begin{tabular}{c|c|c|c|c}
		\hline
		\textbf{Methods} & \textbf{backbone} & \textbf{\makecell{Trainable \\ Parameters}} &  \textbf{Avg. AUC} & \textbf{Avg. OA} \\ \hline
		\textbf{GLFF \cite{ju2023glff}} & ResNet50 & 26.8M & 0.813 & - \\ \hline
		\textbf{UnivFD \cite{ojha2023towards}} & Fixed CLIP & 769 & 0.891 & 0.677 \\ \hline
		\textbf{Full (Ours)} & Fixed CLIP & \textbf{768} & \textbf{0.911} & \textbf{0.774} \\
		\textbf{w/o KGP} & Fixed CLIP & \textbf{768}& 0.884 & 0.754 \\
		\textbf{w/o TTP} & Fixed CLIP & \textbf{768} & 0.903 & 0.598 \\ \hline
	\end{tabular}
\end{table}

\subsection{Comparison with State-of-the-art Methods}
We demonstrate results of deepfak facial image detection with 4 methods learned on spatial domain, i.e., CNN-aug \cite{wang2020cnn}, Nodown \cite{gragnaniello2021gan}, PSM \cite{ju2022fusing}, GLFF \cite{ju2023glff}, and 2 approaches learned on frequency domain, i.e., GAN-DCT \cite{frank2020leveraging}, BeyondtheSpectrum \cite{he2021beyond}. From Table \ref{aucsota} and Table \ref{oasota}, the absolute performance improvement, i.e., $0.098$ on AUC, show the superiority of our proposed method. We achieve the top 4 scores and 1 second-place score out of 6 subsets. For unprocessed subset, Nodown \cite{gragnaniello2021gan} obtains the highest AUC, which argues that down-sampling operation could remove the forgery-related artifacts and lead to a performance decrease.

Compared to GLFF \cite{ju2023glff}, we observe that AUC drops greatly on face blending subset. This is because the resolution of images in this subset is commonly large than $2560 \times 2560$ while the input resolution of our model is $224 \times 224$, needing down-sampling and resulting in poor detection performance. In contrast, PSM \cite{ju2022fusing} and GLFF \cite{ju2023glff} extract local patches to avoid the loss of forgery-related information during down-sampling, obtaining much higher detection performance. Even adopting down-sampling operation, our method still achieves better AUC than UnivFD (i.e., 0.771 vs. 0.705), demonstrating the superiority of ours among CLIP-based methods.

For common post-processing (e.g., JPEG compression and gaussian blur), anti-forensics (e.g., CW attack \cite{carlini2017towards} and GANPrintR \cite{neves2020ganprintr}), multi-image compression and mixed (e.g., the combination of face blending and common post-processing) subsets, our approach has significant advantage compared to other competitors, remarkably boosting the application of deepfake facial image detection in real-world scenarios.

\subsection{Ablation Studies}

\textbf{Architecture Variants:} We conduct ablation studies to evaluate the effectiveness of our proposed Knowledge-Guided Prompt (KGP) learning and Test-Time Prompt (TTP) tuning, as illustrated in Table \ref{ablation}. Without KGP, the performance decrease both on AUC and OA, demonstrating that KGP significantly benefits detection with the help of prior knowledge extracted from large language model. Without TTP, the performance drops slightly on AUC while it decreases greatly on OA, indicating that TTP can effectively alleviate the large domain shift between training and testing categories with the further adaption of prompts on testing data.

\textbf{Trainable Parameters:} Compared to GLFF \cite{ju2023glff} and UnivFD \cite{ojha2023towards}, our method is trained with the least number of parameters while it achieves the best detection performance, as illustrated in Table \ref{ablation}. Compared to GLFF \cite{ju2023glff} that are fully fine-tuned on ResNet50 \cite{he2016deep}, CLIP-based architectures, including our method and UnivFD \cite{ojha2023towards}, obtain much better performance. It shows the great potential of utilizing fixed CLIP (i.e., pre-trained on large corpus such as YFCC100M \cite{thomee2016yfcc100m}) as backbone to address numerous down-stream tasks. Comared to UnivFD \cite{ojha2023towards} that is trained with comparable parameters (i.e., 769 vs. 768), our approach achieves better performance on both AUC and OA, showing its superiority among CLIP-based methods.

\begin{figure}[t]
	\centering
	\includegraphics[scale=0.43]{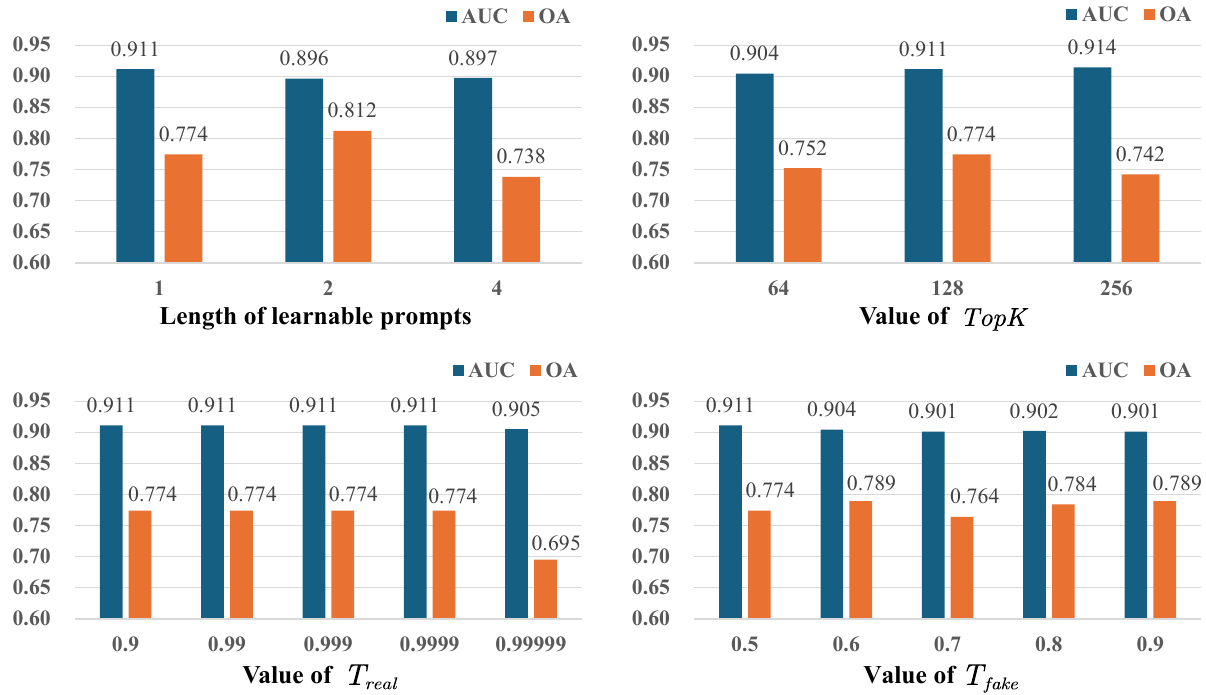}
	\caption{Detection performance with different value of each hyper-parameter.} \label{hyper}
\end{figure}

\textbf{Hyper-parameters:} We conduct detailed experiments with different value of each hyper-parameter and demonstrate the results in Fig. \ref{hyper}. In summary, the performance varies only slightly with changes in the value of each hyper-parameter, especially on the metric of AUC, indicating the robustness of our proposed method to hyper-parameters.

\section{Conclusion}
In this paper, we have proposed knowledge-guided prompt learning and test-time prompt tuning to incorporate prior knowledge from a large language model and alleviate the large domain shift between training categories and testing ones. Our approach achieves notable improvement on deepfake facial image detection performance. They can be seamlessly integrated into other tasks such as anomaly detection, cross domain generalization and so on. In the future, we should devote more efforts on the CLIP-based forgery-related representation learning, without the utilization of down-sampling operations.

\bibliographystyle{IEEEtran}
\bibliography{ref}

\end{document}